\documentclass[11pt]{article}
\usepackage{amsmath}
\usepackage{acl}
\usepackage{booktabs}
\usepackage{times}
\usepackage{latexsym}
\usepackage{booktabs}
\usepackage[table]{xcolor}
\usepackage[T1]{fontenc}
\usepackage[table]{xcolor}
\usepackage[utf8]{inputenc}
\usepackage{booktabs}
\usepackage{longtable}
\usepackage{microtype}
\usepackage{makecell}
\usepackage{inconsolata}

\usepackage{graphicx}

\usepackage{pifont}
\usepackage{xcolor}

\newcommand{\cmark}{\textcolor{green}{\ding{51}}}
\newcommand{\xmark}{\textcolor{red}{\ding{55}}}
\usepackage{acl}
\title{SiMing-Bench: Evaluating Procedural Correctness from Continuous Interactions in Clinical Skill Videos}

\author{
\textbf{Xiyang Huang\textsuperscript{1,2}} \quad
\textbf{Jiawei Lin\textsuperscript{3}} \quad
\textbf{Keying Wu\textsuperscript{1,2}} \quad
\textbf{Jiaxin Huang\textsuperscript{4}} \\
\textbf{Kailai Yang\textsuperscript{5}} \quad
\textbf{Renxiong Wei\textsuperscript{6}} \quad
\textbf{Cheng Zeng\textsuperscript{1}} \quad
\textbf{Jiayi Xiang\textsuperscript{1}} \\
\textbf{Ziyan Kuang\textsuperscript{1,2}} \quad
\textbf{Min Peng\textsuperscript{1,2,$\dagger$}} \quad
\textbf{Qianqian Xie\textsuperscript{1,2,$\dagger$}} \quad
\textbf{Sophia Ananiadou\textsuperscript{5}} \\
\\
\textsuperscript{1}School of Artificial Intelligence, Wuhan University \\
\textsuperscript{2}Center for Language and Information Research, Wuhan University \\
\textsuperscript{3}Southwest Jiaotong University \quad
\textsuperscript{4}MBZUAI \quad
\textsuperscript{5}The University of Manchester \\
\textsuperscript{6}Zhongnan Hospital of Wuhan University \\
}

\begin{document}
\maketitle

\begin{abstract}
Current video benchmarks for multimodal large language models (MLLMs) focus on event recognition, temporal ordering, and long-context recall, but overlook a harder capability required for expert procedural judgment: tracking how ongoing interactions update the procedural state and thereby determine the correctness of later actions. We introduce SiMing-Bench, the first benchmark for evaluating this capability from full-length clinical skill videos. It targets rubric-grounded process-level judgment of whether interaction-driven state updates preserve procedural correctness across an entire workflow. SiMing-Bench is instantiated with SiMing-Score, a physician-annotated dataset of real clinical skill examination videos spanning cardiopulmonary resuscitation, automated external defibrillator operation, and bag-mask ventilation, each paired with a standardized step-wise rubric and dual-expert labels. Across diverse open- and closed-source MLLMs, we observe consistently weak agreement with physician judgments. Moreover, weak performance on rubric-defined intermediate steps persists even when overall procedure-level correlation appears acceptable, suggesting that coarse global assessment substantially overestimates current models’ procedural judgment ability. Additional analyses with binary step judgment and step-aligned clips indicate that the bottleneck is not merely fine-grained scoring or temporal localization, but modeling how continuous interactions update procedural state over time.
\end{abstract}

\section{Introduction}

\begin{figure*}[t]
  \centering
  \includegraphics[width=\textwidth]{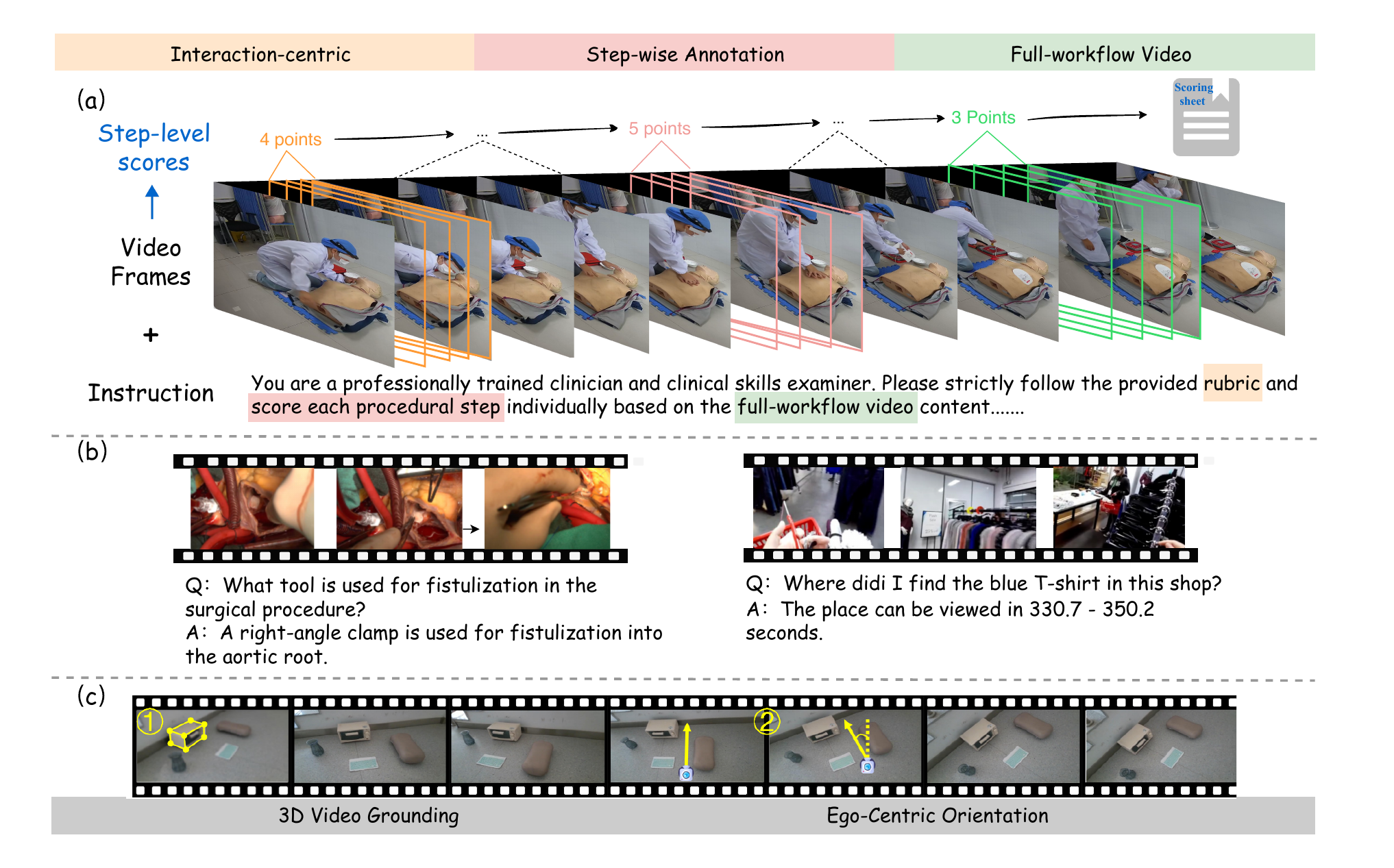}
  \caption{Comparison between SiMing-Bench and prior video benchmarks. (a) SiMing-Bench is built on full-workflow clinical skill videos and paired with expert-defined rubrics and step-wise scores, enabling evaluation of whether models can track continuous interactions throughout an entire procedure and make process-level procedural judgments. (b)(c) Existing benchmarks mainly evaluate localized video understanding, such as instrument recognition, temporal localization, and spatial reasoning.}
  \label{fig:Comparison between existing}
\end{figure*}

Can multimodal large language models (MLLMs) track interaction-driven world evolution in video? Although MLLMs are increasingly studied as world models~\citep{ge2024worldgpt,sreeram2025probing}, it remains unclear whether they can follow how interactions drive a scene from one state to the next over time. Evaluating this ability requires testing whether a model can infer how each action updates the current scene and constrains what can happen next~\citep{chi2025wow}. Clinical skill assessment\footnote{More details in Appendix~\ref{appendix:Clinical Skill Assessment}} is where this ability becomes directly judgeable. In this task, expert examiners determine from video whether a trainee is performing a procedure correctly by interpreting how interactions with instruments, the patient, and the surrounding environment unfold over time. In suturing, for example, a stitching motion may look correct in isolation, yet an examiner can still rule the procedure invalid if the wrong instrument was used or the steps occurred out of order. At the scale of medical training and licensing, automating even part of this expert judgment could remove millions of hours of manual review while turning the same judgments into scalable, fine-grained feedback for clinical training.

However, current MLLM evaluations still primarily assess perception, event understanding, or temporal organization in video, rather than whether unfolding interactions continuously update the procedural state in ways that determine later-step validity (as shown in Table \ref{fig:Comparison between existing}). General video benchmarks such as E.T.Bench~\citep{liu2024bench} and MVBench~\citep{li2024mvbench} evaluate what happened, in what order, and over what duration, but do not directly test whether an earlier interaction changes the conditions under which a subsequent step becomes correct, feasible, or clinically valid. In medical video understanding, benchmarks such as Surgpub-video~\citep{li2025surgpub} and related SurgVLM~\citep{zeng2025surgvlm} settings extend evaluation to domain-specific perception, yet still focus mainly on localized findings or instrument-level recognition, such as which tool appears or whether a lesion is visible, rather than on whether the evolving interaction history preserves procedural correctness as the tool, patient, and operative scene co-develop over time. Spatial reasoning benchmarks~\citep{li2025viewspatial,yang2025mmsi} such as ViewSpatial and MMSI further test positions, viewpoints, and 3D structure, but not whether one action-induced scene change constrains the validity of the next action in an ongoing procedure.

\begin{table*}[t]
\centering
\small
\setlength{\tabcolsep}{5pt}
\begin{tabular}{lcccccc}
\toprule
Benchmark & Domain & \makecell{IC} & \makecell{FV} & \makecell{SA} & \makecell{IPJ} & Major Evaluation Target \\
\midrule
MVBench~\citep{li2024mvbench}        & General Video        & \xmark & \xmark & \xmark & \xmark & Temporal Understanding \\
Video-MME~\citep{fu2025video}      & General Video        & \xmark & \xmark & \xmark & \xmark & Compositional Video Understanding \\
LVBench~\citep{wang2025lvbench}        & General Video        & \xmark & \xmark & \xmark & \xmark & Long Video Understanding \\
EgoTempo~\citep{plizzari2025omnia}       & Egocentric Video     & \xmark & \xmark & \xmark & \xmark & Egocentric Temporal Understanding \\
OmniMMI~\citep{wang2025omnimmi}        & Streaming Video      & \cmark & \xmark & \xmark & \xmark & Streaming Video Understanding \\
ScaleLong~\citep{ma2025scalelong}      & Long Video           & \xmark & \cmark & \xmark & \xmark & Long-video Understanding \\
SurgVLM~\citep{zeng2025surgvlm}  & Medical Video        & \xmark & \xmark & \xmark & \xmark & Surgical Video Understanding \\
Surgpub-video~\citep{li2025surgpub}  & Medical Video        & \cmark & \xmark & \xmark & \xmark & Surgical Video Understanding \\
\midrule
SiMing-Bench   & Clinical Skill Video & \cmark & \cmark & \cmark & \cmark & \textbf{Interaction-driven State Validity Judgment} \\
\bottomrule
\end{tabular}
\caption{Comparison between SiMing-Bench and prior video benchmarks. \textbf{IC} denotes interaction-centric design, \textbf{FV} denotes full-workflow video, \textbf{SA} denotes step-wise annotation, and \textbf{IPJ} denotes interaction-driven procedural judgment. Together, these properties distinguish SiMing-Bench from prior benchmarks by enabling direct evaluation of interaction-driven state validity judgment in long, interaction-rich clinical workflows.}
\label{tab:benchmark_comparison}
\end{table*}

To address this gap, we introduce SiMing-Bench, the first benchmark built on clinical skill videos to test whether MLLMs can judge procedural correctness from continuous interactions in real-world environments. It requires models to track interaction-driven state updates and make rubric-grounded procedural judgments over an entire workflow. SiMing-Bench is built on SiMing-Score, a manually annotated dataset of 200 full-length clinical skill videos spanning three skill categories, with each 2 to 4 minute video preserving an entire workflow within a single scene. As shown in Figure \ref{fig:Comparison between existing} (a), collected from medical student examinations at three hospitals, the videos capture sustained interactions among the trainee, medical instruments, the patient, and the surrounding environment over time. Two physicians independently scored each video using standardized assessment forms, yielding step-level textual scoring criteria, ordinal ratings for multiple procedural steps, and an overall performance score. Given a full video and its scoring rubric, the task is to predict a discrete score for each predefined step as well as an overall score on the same scale used by human raters. Evaluation is conducted at two levels, with Pearson correlation(PLCC) and Spearman correlation(SRCC) measuring overall-score consistency and quadratically weighted Cohen’s $\kappa$ measuring step-level consistency.
 
Across 12 MLLMs with reported main results, we observe a persistent gap between model judgment and expert assessment. On overall procedure scoring, the strongest model, GPT-4o, achieves only 0.158 PLCC, while still yielding zero step-level weighted Cohen’s $\kappa$. In contrast, physician-physician agreement reaches 0.871 PLCC on overall scores and 0.732 Cohen’s $\kappa$ on step-level labels. This large gap shows that current MLLMs still struggle to convert continuous interactions in clinical videos into reliable procedural judgment. Moreover, weak overall-score correlation can coexist with near-zero agreement on rubric-defined intermediate steps, indicating that coarse global assessment substantially overestimates current models’ process-level understanding. Additional diagnostic analyses further show that this limitation is not resolved by simplifying the task. Under binary correctness judgment, the best model reaches only 0.132 MCC. Under manually segmented step-aligned clips, the best step-level $\kappa$ reaches only 0.075. Together, these results show that current MLLMs remain limited as world models because they do not yet reliably turn evolving interaction states into explicit judgments about procedural validity.

In summary, our main contributions are threefold.
\textbf{First}, we introduce \textbf{SiMing-Bench}, the first benchmark built on clinical skill videos to test whether MLLMs can judge procedural correctness from continuous interactions. It targets a missing capability in current video evaluation, namely process-level judgment over long, interaction-rich procedures.
\textbf{Second}, we formulate this problem as a multi-granularity assessment task and construct \textbf{SiMing-Score}, a manually annotated dataset of 200 full-length videos across three clinical skill categories. Each video is independently scored by two physicians with step-level scoring criteria, ordinal ratings, and an overall score.
\textbf{Third}, experiments on diverse MLLMs reveal a persistent gap between model scoring and expert judgment. Further analyses show that this gap is not resolved by simplifying score granularity or temporal localization, suggesting a deeper limitation in how current MLLMs model interaction-driven procedural state.

\section{Related Work}
\subsection{Video Understanding Benchmarks}
Recent benchmarks have substantially expanded the evaluation of MLLMs for video understanding, covering event comprehension, temporal reasoning, long-context modeling, streaming perception, and open-ended multimodal understanding. Early comprehensive benchmarks such as MVBench~\citep{li2024mvbench} and Video-MME~\citep{fu2025video} mainly assess event recognition and multimodal comprehension across diverse video domains, while more recent datasets such as LVBench~\citep{wang2025lvbench} and EgoTempo~\citep{plizzari2025omnia} further emphasize long-video understanding and fine-grained temporal reasoning. Recent efforts such as OmniMMI~\citep{wang2025omnimmi} and ScaleLong~\citep{ma2025scalelong} extend evaluation to streaming and hierarchical settings, testing whether models can perceive, organize, and retain information across long and continuously unfolding videos. In parallel, domain-specific medical video benchmarks, including SurgVLM~\citep{zeng2025surgvlm} and Surgpub-video~\citep{li2025surgpub}, have moved beyond frame-level recognition toward broader surgical intelligence. However, as illustrated in Table~\ref{tab:benchmark_comparison}, existing benchmarks mainly focus on event recognition, temporal reasoning, or long-context understanding, rather than on whether continuous interactions progressively update scene states in ways that determine procedural correctness in subsequent steps throughout long workflows.

\subsection{MLLMs for Video Understanding}
Recent video MLLMs have increasingly evolved from video-adapted extensions toward unified multimodal models with stronger long-context and spatiotemporal reasoning capabilities. Models such as LLaVA-Video~\citep{zhang2024llava} and LLaVA-OneVision~\citep{li2024llava} extend general-purpose MLLMs to open-ended video understanding, while InternVL3.5~\citep{wang2025internvl3} and Qwen3-VL~\citep{bai2025qwen3} further strengthen long-context multimodal perception and spatiotemporal reasoning. More recent models, including Molmo2~\citep{clark2026molmo2}, TimeSuite~\citep{zeng2024timesuite}, LLaVA-MR~\citep{lu2024llava}, and ReMoRa~\citep{yashima2026remora}, also improve grounding, temporal localization, and long-video modeling. Despite these advances, existing video MLLMs are still primarily optimized to understand what happened in a video, where it occurred, and how events unfold over time. They are less explicitly designed to judge whether continuous interactions progressively update the scene state in ways that determine the correctness, feasibility, or validity of subsequent actions in a long procedure. This gap is closely related to recent vision-language world modeling studies~\citep{ge2024worldgpt,chen2025planning}, which move beyond event recognition toward modeling state transitions, action consequences, and temporally structured future evolution. However, it remains unclear whether current MLLMs can robustly track interaction-driven world-state updates for fine-grained procedural judgment, or instead rely on superficial sequential regularities in observed state changes.

\section{SiMing-Bench}
SiMing-Bench studies whether an MLLM can assess \emph{procedural correctness} from a complete clinical skill video under an expert-defined scoring rubric. Unlike conventional video understanding tasks that ask what happened in a clip, the task requires mapping a long-form procedure to rubric-aligned judgments over clinically meaningful steps. Each step corresponds to a temporally extended interaction segment rather than a single visual event, and its correctness depends on whether the preceding interactions have established the procedural conditions required for the current action. The model must therefore evaluate the procedure as an evolving interaction process, rather than as a collection of isolated observations.
\subsection{Task Definition}
We formulate this setting as a structured prediction problem over long-form videos. Given a complete clinical skill video $V=\{V_t\}_{t=1}^{T}$, and a step-wise rubric $R=\{r_i\}_{i=1}^{N}$, the task requires a model $F_\theta$ predicting a sequence of step-level scores:
\[
\hat{\mathbf{s}}=\{\hat{s}_i\}_{i=1}^{N}=F_\theta(V,R).
\]
where $V_t$ denotes the visual observation at time step $t$, $r_i$ specifies the scoring criteria and point rules for step $i$. Here, each $\hat{s}_i \in \mathcal{Y}_i$ is a discrete score for step $i$, and $\mathcal{Y}_i$ is the valid score set defined by the rubric for that step. The overall procedure score is obtained by aggregating the predicted step-level scores:
\[
\hat{y}=\sum_{i=1}^{N}\hat{s}_i,
\]
where $\hat{y}$ denotes the predicted total score for the full procedure.

\subsection{Dataset Curation}

\begin{table}[t]
\centering
\small
\setlength{\tabcolsep}{6pt}
\begin{tabular}{lcc}
\toprule
Skill & Avg. Duration & \#Rubric Steps \\
\midrule
CPR & 85.6 s & 19 \\
AED & 50.2 s & 9 \\
BMV & 33.1 s & 5 \\
\midrule
Total videos & \multicolumn{2}{c}{200} \\
Total duration & \multicolumn{2}{c}{563 min} \\
Avg. duration per sample & \multicolumn{2}{c}{168.9 s} \\
Total rubric steps & \multicolumn{2}{c}{33} \\
\bottomrule
\end{tabular}
\caption{Dataset statistics of SiMing-Bench. Each video contains a complete assessment including CPR, AED, and BMV, rather than a single isolated skill. We report the average duration and rubric steps for each skill, together with the overall dataset scale after quality control.}
\label{tab:dataset_statistics}
\end{table}

\textbf{Data source and collection.}
SiMing-Score is curated from real clinical skill assessment videos involving medical students at three hospitals, forming, to our knowledge, the first benchmark of full-length clinical skill videos for rubric-grounded procedural assessment. The videos were collected in standardized clinical skill training and examination settings, where students performed complete procedures under expert-defined assessment protocols rather than producing expert demonstrations.
We collected 233 videos in total. Each video includes assessments of cardiopulmonary resuscitation (CPR), automated external defibrillator (AED) operation, and bag-mask ventilation (BMV). The average durations of the three skills are 85.6s, 50.2s, and 33.1s, respectively. After quality control, the dataset contains approximately 563 minutes of video, with an average duration of 168.9s per sample. Detailed dataset statistics are summarized in Table~\ref{tab:dataset_statistics}. Each video records one complete attempt within a single testing scene, preserving the full procedural trajectory from start to finish rather than isolated clips or manually segmented events. The data collection and research use protocols were reviewed and approved by the Institutional Review Board (IRB)\footnote{More details in Appendix~\ref{Appendix_Ethical}} of the participating institution.

Because the performers are students in authentic assessment settings, the videos naturally exhibit clinically meaningful variation in procedural quality, including incorrect ordering, incomplete execution, unstable coordination, and other deviations that directly affect rubric-based scoring. Procedural assessment becomes meaningful precisely because correctness is not assumed, but must be judged from the full interaction process. The three skill categories instantiate complementary forms of long-horizon procedural interaction\footnote{The Appendix~\ref{Appendix A.1} provides detailed descriptions of CPR, AED operation, and BMV, including their procedural characteristics and representative action sequences.}. 

For each skill, the scoring rubric specifies clinically meaningful sub-steps together with explicit requirements for correct execution, including action order, continuity, device handling, safety constraints, and coordination quality. All recordings were obtained under standardized examination conditions, with fixed examination protocols, comparable room layouts, and clear visibility of rubric-relevant interactions among the trainee, patient simulator, and surrounding instruments.

\paragraph{Physician Rubrics and Dual-Expert Annotation.} Each skill category is paired with a standardized step-wise scoring rubric used in real clinical skill examinations, with full rubric details provided in the Appendix~\ref{Appendix_Rubric}. These rubrics are developed by physicians for procedural assessment and decompose each clinical task into predefined sub-steps with explicit scoring criteria. The resulting labels therefore reflect expert judgment of procedural correctness rather than free-form description or proxy supervision.

Using these rubrics, two clinically trained physician annotators\footnote{More details in Appendix~\ref{appendix:Privacy and Annotation}} independently review each full-length video and assign scores to every predefined step. Each annotator evaluates the complete procedure from start to finish and determines, for each step, whether the trainee satisfies the rubric-defined requirements for correct execution. The overall procedure score is then obtained by aggregating the expert-assigned step-level scores under the rubric-defined scoring scheme. Each sample thus contains dual-expert, rubric-aligned step annotations together with a total score grounded in the full interaction process.

\paragraph{Quality control and adjudication.} Quality control was applied at both the video curation and annotation stages. At the data level, we retained only videos with complete procedural coverage and clear visibility of rubric-relevant interactions. Videos were excluded only when the recording itself was insufficient for rubric-based assessment, such as when key stages were missing because of interruption, severe occlusion, or incomplete capture. In contrast, procedurally incorrect or low-quality student performances were retained, since such cases are intrinsic to real clinical skill assessment and are essential for evaluating rubric-grounded judgments of procedural correctness.

At the annotation level, we used a two-stage protocol. In the first stage, two physician annotators independently reviewed each full-length video and assigned step-level scores according to the official rubric. In the second stage, samples with discrepant step-level labels were jointly re-examined against the rubric, and disagreements were resolved step by step through adjudication. The final consensus step-level labels were then used to derive the overall procedure score. Inter-annotator agreement before adjudication is reported in Table~\ref{tab:agreement}. In total, 65 step instances(1.06\%) across 31 videos(15.5\%) required adjudication, and 33 videos were excluded during curation because the recording was not sufficient for rubric-based assessment.

\begin{table}[t]
\centering
\small
\begin{tabular}{lc}
\toprule
\textbf{Metric} & \textbf{Value} \\
\midrule
Overall-score agreement (PLCC) & 0.871 \\
Step-level agreement (Cohen'$\kappa$) & 0.732 \\
Step instances requiring adjudication & 65 (1.06\%) \\
Videos excluded during curation & 33 \\
\bottomrule
\end{tabular}
\caption{Annotation agreement and curation statistics before adjudication.}
\label{tab:agreement}
\end{table}

\begin{table*}[t]
\centering
\small
\setlength{\tabcolsep}{4pt}
\resizebox{\textwidth}{!}{%
\begin{tabular}{lcccccccc}
\toprule
Models & Parameters & PLCC & SRCC & Cohen's $\kappa$ & MCC & P & R & $F_1$ \\
\midrule
\multicolumn{9}{l}{\textcolor[HTML]{9B9B9B}{\textit{Open-source Video-MLLMs}}} \\
\midrule
LLaVA-Video~\citep{zhang2024llava}      & 7B  & 0.104  & -0.095 & 0.033  & -0.033 & 0.157 & 0.135 & 0.145 \\
LLaVA-Video~\citep{zhang2024llava}      & 72B & -0.047 & -0.015 & 0.044  & 0.039  & 0.207 & 0.369 & 0.266 \\
LLaVA-OneVision~\citep{li2024llava}     & 7B  & 0.038  & 0.123  & 0.038  & 0.029  & 0.145 & 0.059 & 0.084 \\
Qwen3-VL~\citep{bai2025qwen3}           & 30B & 0.037  & 0.099  & -0.010 & 0.0434 & 0.174 & 0.659 & 0.276 \\
\midrule
\multicolumn{9}{l}{\textcolor[HTML]{9B9B9B}{\textit{Omni-MLLMs}}} \\
\midrule
Qwen3-Omni~\citep{xu2025qwen3}          & 30B  & -0.002 & 0.011  & 0.000   & 0.000   & 0.183 & 1.000     & 0.310 \\
GPT-4o~\citep{openai2024gpt4o}          & --   & 0.158  & 0.169  & 0.000      & -0.023 & 0.156 & 0.070 & 0.097 \\
\midrule
\multicolumn{9}{l}{\textcolor[HTML]{9B9B9B}{\textit{Medical Video-MLLMs}}} \\
\midrule
UniMed-VL~\citep{ning2025unimedvlunifyingmedicalmultimodal} & 14B & -0.035 & -0.053 & -0.006 & 0.053  & 0.257 & 0.107 & 0.152 \\
MedGemma~\citep{sellergren2025medgemma}                     & 27B & -0.059 & -0.074 & 0.007  & 0.028  & 0.201 & 0.393 & 0.266 \\
MediX-R1~\citep{mullappilly2026medix}                       & 30B & -0.046 & -0.058 & -0.048 & -0.039 & 0.176 & 0.461 & 0.254 \\
\midrule
\multicolumn{9}{l}{\textcolor[HTML]{9B9B9B}{\textit{Commercial MLLMs}}} \\
\midrule
Doubao-1.6-Vision~\citep{volcengine2025doubaoseed16vision} & 230B & -0.055 & -0.034 & 0.037 & -0.031 & 0.150 & 0.081 & 0.105 \\
Gemini-3 Pro~\citep{google2025gemini3}                     & --   & 0.022  & -0.116 & 0.018 & -0.020 & 0.157 & 0.106 & 0.126 \\
GPT-5.2~\citep{openai2025gpt52}                            & --   & -0.053 & 0.016  & 0.048 & 0.132  & 0.212 & 0.909 & 0.344 \\
\bottomrule
\end{tabular}%
}
\caption{Main results on SiMing-Bench. We evaluate diverse MLLMs under the rubric-grounded full-video setting at both the overall-score and step levels. MCC, Precision, Recall, and $F_1$ are additionally reported for binary step correctness judgment (more details in Section~\ref{sec:Diagnostic Analysis of Step-level Failure}).}
\label{tab:main_results}
\end{table*}

\begin{table}[t]
\centering
\scriptsize
\setlength{\tabcolsep}{4pt}
\begin{tabular}{lccc}
\toprule
Models & PLCC & SRCC & Cohen's $\kappa$ \\
\midrule
LLaVA-Video-7B
& -0.150 {\scriptsize (-0.254)}
& -0.170 {\scriptsize (-0.075)}
& 0.009 {\scriptsize (-0.024)} \\
LLaVA-Video-72B
& \colorbox[HTML]{F8CECC}{0.057} {\scriptsize (+0.104)}
& \colorbox[HTML]{F8CECC}{0.098} {\scriptsize (+0.113)}
& -0.003 {\scriptsize (-0.047)} \\
LLaVA-OneVision
& 0.000 {\scriptsize (-0.038)}
& 0.000 {\scriptsize (-0.123)}
& 0.000 {\scriptsize (-0.038)} \\
Qwen3-VL
& \colorbox[HTML]{F8CECC}{0.135} {\scriptsize (+0.098)}
& \colorbox[HTML]{F8CECC}{0.125} {\scriptsize (+0.026)}
& \colorbox[HTML]{F8CECC}{0.000} {\scriptsize (+0.010)} \\
Qwen3-Omni
& \colorbox[HTML]{F8CECC}{0.092} {\scriptsize (+0.094)}
& \colorbox[HTML]{F8CECC}{0.110} {\scriptsize (+0.099)}
& \colorbox[HTML]{F8CECC}{0.075} {\scriptsize (+0.075)} \\
GPT-4o
& -0.018 {\scriptsize (-0.176)}
& 0.004 {\scriptsize (-0.165)}
& \colorbox[HTML]{F8CECC}{0.019} {\scriptsize (+0.019)} \\
UniMed-VL
& \colorbox[HTML]{F8CECC}{0.084} {\scriptsize (+0.119)}
& \colorbox[HTML]{F8CECC}{0.061} {\scriptsize (+0.114)}
& \colorbox[HTML]{F8CECC}{0.016} {\scriptsize (+0.022)} \\
MedGemma
& \colorbox[HTML]{F8CECC}{0.030} {\scriptsize (+0.089)}
& \colorbox[HTML]{F8CECC}{0.079} {\scriptsize (+0.153)}
& 0.002 {\scriptsize (-0.005)} \\
MediX-R1
& \colorbox[HTML]{F8CECC}{0.108} {\scriptsize (+0.154)}
& \colorbox[HTML]{F8CECC}{0.121} {\scriptsize (+0.179)}
& \colorbox[HTML]{F8CECC}{0.010} {\scriptsize (+0.058)} \\
Doubao-1.6-Vision
& \colorbox[HTML]{F8CECC}{0.055} {\scriptsize (+0.110)}
& \colorbox[HTML]{F8CECC}{0.032} {\scriptsize (+0.066)}
& 0.006 {\scriptsize (-0.031)} \\
Gemini-3 Pro
& \colorbox[HTML]{F8CECC}{0.331} {\scriptsize (+0.447)}  
& \colorbox[HTML]{F8CECC}{0.296} {\scriptsize (+0.412)} 
& 0.015(-0.003)  \\
GPT-5.2
& \colorbox[HTML]{F8CECC}{0.147} {\scriptsize (+0.200)}
& \colorbox[HTML]{F8CECC}{0.157} {\scriptsize (+0.141)}
& 0.011 {\scriptsize (-0.037)} \\
\bottomrule
\end{tabular}
\caption{Comparison between the full-video setting and the step-aligned clips setting on PLCC, SRCC and Cohen's $\kappa$. Improvements are highlighted with \colorbox[HTML]{F8CECC}{\strut red} backgrounds.}
\label{tab:5}
\end{table}

\begin{figure}[t]
    \centering
    \includegraphics[width=\linewidth]{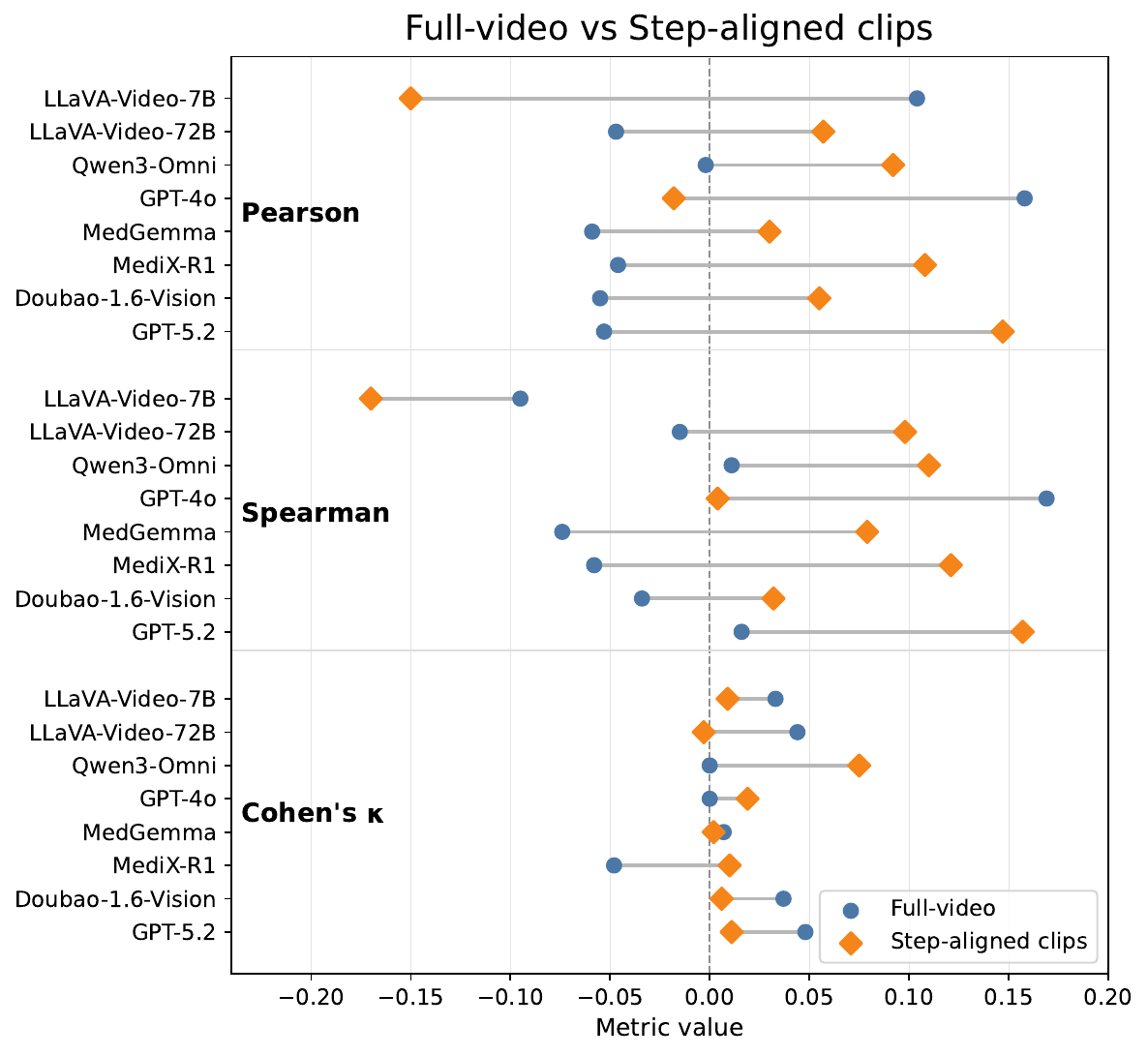}
    \caption{Comparison between the full-video setting and the step-aligned clip setting. Each dumbbell connects the same model under the two evaluation settings.}
    \label{fig:dumbbell}
\end{figure}
\subsection{Evaluation}
We evaluate MLLMs on SiMing-Bench under a rubric-grounded prompting protocol, where each structured clinical scoring rubric is converted into a textual prompt and the model predicts step-level scores for the full procedure. Because clinical skill assessment requires both global judgment over the entire procedure and local judgment over clinically meaningful sub-steps, we evaluate model performance at two complementary levels.\\
\textbf{Overall-score consistency.}
We first measure whether model predictions agree with expert judgments on the quality of the complete procedure. Let $\hat{y}$ and $y$ denote the predicted and expert overall scores, respectively. We report Pearson correlation(PLCC) and Spearman correlation(SRCC) over the evaluation set. PLCC measures agreement in absolute scoring tendency, while SRCC measures whether the model preserves the expert ranking of procedural quality across samples. The formulations are provided in Appendix~\ref{Appendix_Metric_Overall-score}.\\
\textbf{Step-level consistency.}
We further measure whether the model agrees with experts on the correctness of individual procedural steps. Since step-level ratings are ordinal rather than interval-scaled, we evaluate agreement using quadratic weighted Cohen's $\kappa$. Both expert and predicted step scores are mapped to four ordered categories, corresponding to fully correct, basically correct with minor issues, clearly deficient, and severely incorrect or failed. This metric penalizes larger disagreements more heavily and is therefore appropriate for rubric-based procedural assessment. Implementation details are provided in Appendix~\ref{Appendix_Metric_Step-score}.

Unless otherwise specified, overall-score metrics are computed against the expert consensus score, and step-level agreement is computed over all rubric-defined steps in the evaluation set.

\section{Experiments}
\subsection{Experimental Settings}
\textbf{Baseline models.} To comprehensively evaluate model performance on SiMing-Bench, we compare a diverse set of representative baselines spanning general-purpose open-source Video-LLMs~\citep{zhang2024llava,li2024llava,bai2025qwen3}, omni-modal large models~\citep{xu2025qwen3,openai2024gpt4o}, medical-domain MLLMs~\citep{ning2025unimedvlunifyingmedicalmultimodal,sellergren2025medgemma,mullappilly2026medix}, and commercial proprietary MLLMs~\citep{volcengine2025doubaoseed16vision,google2025gemini3,openai2025gpt52}. This selection allows us to assess procedural understanding ability across models with different architectures, training data, and domain specialization. 

\subsection{Main Results Analysis}

\textbf{Current MLLMs fail to achieve expert-aligned process-level judgment. }
Table~\ref{tab:main_results} shows that all evaluated models perform poorly on SiMing-Bench, with uniformly weak agreement with expert ratings at both the overall-score and step levels. Even GPT-4o, the strongest model on overall-score evaluation, reaches only 0.158 PLCC and 0.169 SRCC, while yielding zero step-level $\kappa$. Among open-source models, LLaVA-OneVision achieves the highest SRCC, but no model produces consistent gains across evaluation levels. The dominant pattern is therefore not isolated weakness in particular systems, but a systematic failure to convert long-horizon interaction history into expert-aligned, rubric-grounded process-level judgment. This failure also does not diminish reliably with scale: performance does not improve monotonically, and both open- and closed-source systems remain well below expert agreement under zero-shot evaluation.

\textbf{Overall-score agreement masks failures in step-level procedural assessment. }
A central finding is the persistent mismatch between overall-score consistency and step-level consistency. GPT-4o, for example, achieves the strongest correlation with expert overall scores, yet fails to obtain positive agreement on rubric-defined steps. Similar discrepancies appear across other models. This indicates that weak agreement at the full-procedure level can arise without correct judgment of the intermediate interactions that determine procedural validity. In other words, current models may capture coarse signals about global procedure quality while still failing to identify which earlier interactions changed the procedural state and why later actions became correct or incorrect. This result directly supports the multi-granularity design of SiMing-Bench: evaluating only the overall score would substantially overestimate current models' ability to perform expert-like process-level assessment.

\textbf{Neither scale nor medical specialization resolves the underlying failure. } 
Increasing model scale does not reliably improve performance on SiMing-Bench. For example, LLaVA-Video-72B underperforms its 7B counterpart on PLCC, and larger commercial models also remain near zero in agreement. Medical-domain models likewise fail to show a consistent advantage. UniMed-VL, MedGemma, and MediX-R1 remain near or below zero on the reported metrics, indicating that medical specialization alone is insufficient for rubric-grounded procedural assessment. This is consistent with the task design of SiMing-Bench, which does not primarily test recognition of medical objects or terminology, but whether a model can represent how continuous interactions update procedural conditions and determine the validity of subsequent actions.

SiMing-Bench therefore exposes a concrete world-modeling bottleneck in current video MLLMs: they do not reliably infer how earlier interactions reshape the conditions under which later actions become correct, incorrect, feasible, or invalid.

\subsection{Diagnostic Analysis of Step-level Failure}
\label{sec:Diagnostic Analysis of Step-level Failure}
To diagnose the source of step-level failure, we conduct two additional analyses that separately reduce the difficulty of score assignment and temporal evidence search.

\begin{figure*}[t]
    \centering
    \includegraphics[width=0.9\linewidth]{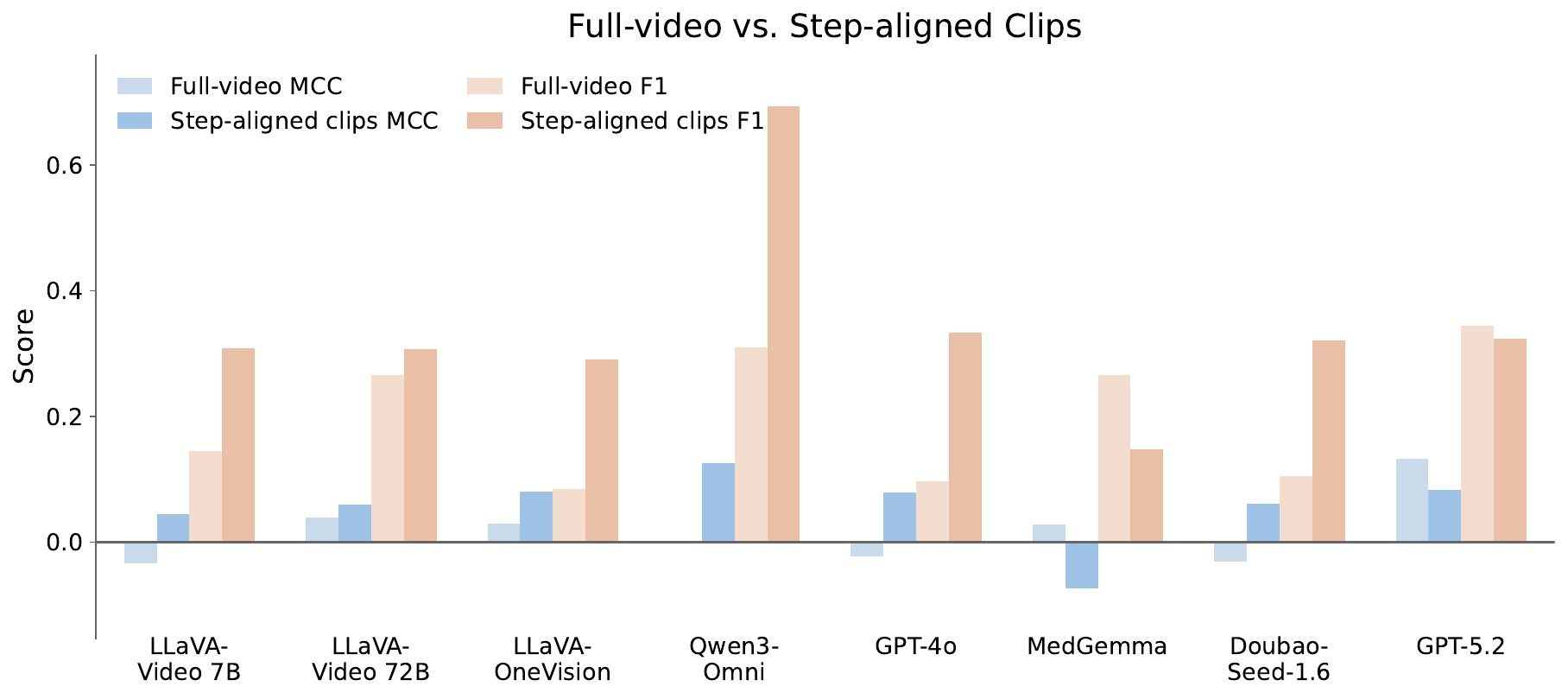}
    \caption{Comparison between the full-video setting and the step-aligned clips setting on binary step judgment. For each model, we report MCC and F1 under both settings.}
    \label{fig:bar}
\end{figure*}

\textbf{Binary correctness judgment does not resolve the failure.}
We first simplify step-level evaluation from ordinal scoring to binary correctness classification, where the model only decides whether each interaction step is correctly performed. This removes the need to map visual evidence to fine-grained score levels and isolates a more basic capability, judging whether a clinically meaningful interaction satisfies the procedural requirement. Performance nevertheless remains weak across all metrics. As shown in Table~\ref{tab:main_results}, most models achieve MCC values close to zero, and some are negative; even the best model, GPT-5.2, reaches only 0.132 MCC. Moreover, some seemingly stronger F1 scores are driven by heavily biased prediction behavior rather than reliable step discrimination. For example, Qwen3-Omni attains perfect recall but extremely low precision and zero MCC, indicating a strong tendency to predict most steps as correct. This suggests that the bottleneck is not merely fine-grained score assignment, but the underlying ability to distinguish correct from incorrect interactions.

\textbf{Step-aligned temporal localization yields only limited gains.}
We next evaluate models on manually segmented step-aligned clips, where each input contains only a short clip and the corresponding step requirement. This removes the need to search over the full procedure for relevant evidence. Figure~\ref{fig:dumbbell} illustrates that while some models show modest improvement in overall-score correlations, step-level agreement remains extremely weak. As shown in Table~\ref{tab:5}, $\kappa$ stays near zero for almost all models, with the best result reaching only 0.075. Binary correctness evaluation shows a similar pattern. Figure~\ref{fig:bar} shows that, although step segmentation improves performance for some models, the gains remain limited and often coexist with strong positive prediction bias. These results indicate that long-horizon temporal search is only part of the challenge. Even when relevant evidence is pre-localized, current MLLMs still struggle to judge whether the observed interaction satisfies the procedural criterion.

Taken together, these diagnostic analyses show that the main bottleneck is neither fine-grained score mapping alone nor temporal localization alone. Instead, current MLLMs fail at the more fundamental problem of converting observed interactions into expert-aligned procedural judgment, consistent with a broader failure to maintain interaction-updated procedural state over the course of a procedure.

\section{Conclusion}
We introduce SiMing-Bench, a benchmark for evaluating expert-aligned process-level judgment from full-length clinical skill videos. Beyond event recognition or long-context recall, SiMing-Bench tests whether MLLMs can model how ongoing interactions update procedural conditions and thereby determine the correctness of later actions. Across overall-score, step-level, binary judgment, and step-aligned clip settings, current MLLMs consistently show weak agreement with physician experts. Moreover, weak performance on rubric-defined intermediate steps persists even when overall procedure-level correlation appears acceptable, suggesting that coarse agreement substantially overestimates true procedural assessment ability. The limited gains from binary simplification and pre-localized clips further indicate that the core bottleneck is not merely scoring difficulty or temporal evidence search, but interaction-grounded procedural state modeling. We hope SiMing-Bench encourages future multimodal systems to move beyond event recognition toward expert-aligned judgment over continuously evolving real-world processes.

\section{Limitations}

SiMing-Bench focuses on clinical skill assessment in standardized educational examination settings. This design provides high-quality rubric-grounded supervision and controlled procedural comparisons, but it does not cover the full variability of real clinical practice, where patient conditions, environment, team coordination, and unexpected events may be substantially more complex. Our conclusions should therefore be interpreted as evidence of a core failure in process-level procedural judgment under structured assessment, rather than as a complete characterization of video understanding in all clinical environments.

The benchmark currently covers three clinical skill categories, including cardiopulmonary resuscitation, automated external defibrillator operation, and bag-mask ventilation. These tasks were chosen because they involve long-form, interaction-dependent procedures with expert-defined rubrics, but they do not exhaust the space of clinical procedural reasoning. Other settings, such as surgery, bedside examination, or multi-actor team procedures, may involve different temporal dependencies and decision structures.

Our evaluation is also rubric-grounded by design. This is a strength for standardized assessment, but it means that the benchmark emphasizes agreement with expert scoring protocols rather than all possible forms of clinical reasoning. In particular, successful performance on SiMing-Bench would not by itself imply broader competence in diagnosis, treatment planning, or open-ended clinical decision making.

Finally, although we include diagnostic analyses with binary judgment and step-aligned clips, these experiments do not fully disentangle all possible sources of model failure. The observed performance gaps are consistent with a limitation in interaction-grounded procedural state modeling, but future work is still needed to more explicitly probe memory, localization, causal attribution, and latent world-state tracking in long-form multimodal reasoning.

\section{Ethical Considerations}

This work studies MLLMs in the context of clinical skill assessment, a domain in which errors can have educational and safety implications. SiMing-Bench is intended strictly as a research benchmark for evaluating process-level multimodal reasoning. It is not designed or validated for direct deployment in medical education scoring, competency certification, or clinical decision support without substantial additional validation, oversight, and regulatory review.

The dataset is curated from standardized clinical skill training and examination scenarios involving medical students. The data collection and research use protocols were reviewed and approved by the Institutional Review Board (IRB)\footnote{More details in Appendix~\ref{Appendix_Ethical}} of the participating institution.
Because the videos concern educational assessment in a medically themed setting, careful handling of privacy and institutional permissions is essential. Data collection, storage, annotation, and release should comply with the relevant institutional review, consent, and de-identification requirements applicable to the participating sites. Any public release should avoid personally identifying information and should preserve only the information necessary for benchmark evaluation. 

A further ethical risk is overinterpretation of model capability. Since the benchmark measures rubric-grounded procedural judgment, strong performance would not imply that a model is safe for real patient-facing use. Conversely, weak performance should not be misread as a judgment about student competence outside the specific scoring rubric and examination context. We therefore frame all results as model-evaluation findings rather than educational or clinical decisions about individuals.

Finally, the benchmark may contribute to future automated assessment tools, which creates both opportunity and risk. Such systems could support training and feedback, but they could also amplify annotation biases, institutional scoring conventions, or unjustified trust in model outputs. For this reason, any downstream use should preserve human expert oversight, make scoring criteria transparent, and treat model outputs as assistive rather than authoritative.

\bibliographystyle{acl_natbib}
\bibliography{custom}

\clearpage

\appendix
\section{Clinical Skill Assessment}
\label{appendix:Clinical Skill Assessment}

\begin{figure*}[t]
  \centering
  \includegraphics[width=\textwidth]{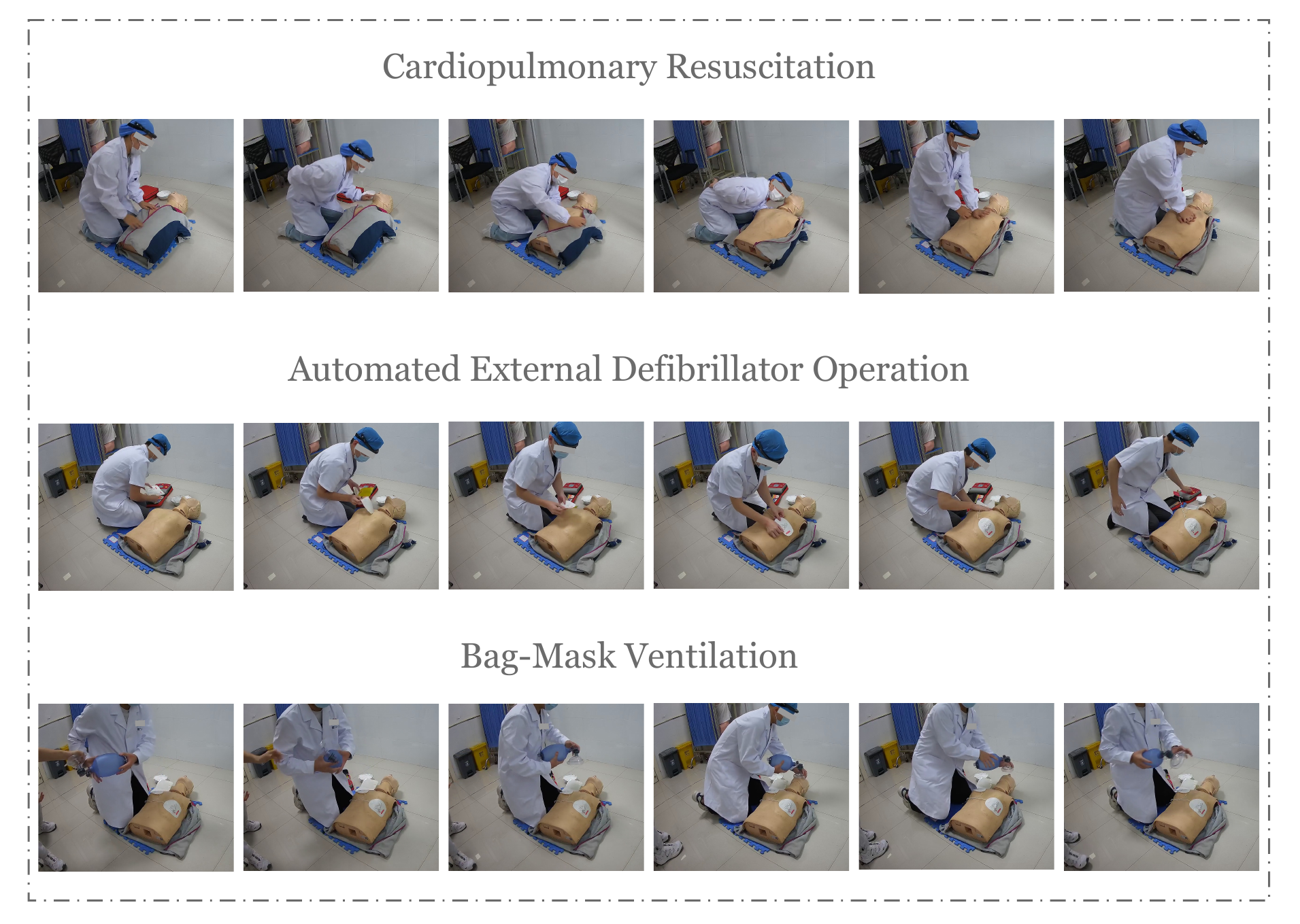}
  \caption{Multi-frame examples of the three emergency care skills considered in this study are shown: cardiopulmonary resuscitation (CPR), automated external defibrillator (AED) operation, and bag-mask ventilation (BMV). The figure illustrates how each procedure unfolds as a temporally ordered sequence of coordinated actions, highlighting the highly structured and interaction-intensive nature of these tasks.}
  \label{append:Fig skills}
\end{figure*}

\subsection{What is clinical skills assessment?}
\label{Appendix A.1}
Clinical skills assessment refers to performance-based approaches in medical education used to evaluate whether learners can integrate communication, knowledge, technical skills, and clinical reasoning in practice-relevant tasks~\citep{epstein2002defining}. Rather than assessing only what learners know, such assessment is commonly understood as targeting higher levels of performance beyond knowledge alone~\citep{miller1990assessment}. One of the most widely used formats for this purpose is the Objective Structured Clinical Examination (OSCE)~\citep{khan2013objective}, which has become widely adopted in both undergraduate and postgraduate clinical education. In an OSCE, learners rotate through a series of stations organized around specific clinical tasks, with predefined scoring approaches and standardized administration procedures intended to improve the structure and objectivity of clinical performance assessment.
In our study, we focus on procedural clinical skills in emergency care, specifically cardiopulmonary resuscitation (CPR), automated external defibrillator (AED) operation, and bag-mask ventilation (BMV). As shown in Figure~\ref{append:Fig skills}, we present multi-frame examples of these skills to illustrate how each procedure unfolds over time through a sequence of coordinated actions. These skills are highly structured, time-sensitive, and interaction-intensive, requiring learners to complete a full sequence of clinically meaningful actions rather than isolated subtasks. This processual nature makes them well suited for evaluating fine-grained procedural competence, including step-wise execution quality, action ordering, and consistency across the entire workflow.

\subsection{Why is it important?}
Clinical skills assessment is important for at least three reasons. First, it is closely tied to patient safety and quality of care, as medical training must ultimately ensure that learners can translate knowledge into safe and effective clinical practice rather than merely reproduce factual information in written examinations~\citep{walton2011republished}. Second, it addresses a fundamental limitation of written tests. As emphasized in the classic framework of clinical competence, written examinations mainly assess whether learners “know” or “know how,” whereas clinical skills assessments are better suited to evaluating whether learners can actually demonstrate performance in practice, namely whether they can “show how” and, to some extent, “do”~\citep{miller1990assessment}. This distinction is particularly important in medical education, where the ultimate goal is not merely knowledge retention, but the ability to function effectively in authentic clinical encounters~\citep{harden1975assessment}. Third, clinical skills assessment supports not only individual learner development but also the continuous improvement of the curriculum and the assessment program itself. Contemporary medical education increasingly views assessment as a programmatic process that informs feedback, guides learning, and provides evidence for improving educational design at the program level~\citep{van2005assessing}.

\subsection{What challenges does it face?}
Clinical skills assessment faces several persistent challenges. First, it is resource-intensive, particularly when implemented in the form of OSCEs. OSCEs are widely recognized as difficult to organize, requiring substantial human and material resources and often involving greater time and cost than traditional or less structured assessment formats~\citep{cusimano1994comparative}. Second, scoring consistency remains limited despite the standardized design of OSCEs. Although OSCEs were developed to improve the structure and objectivity of clinical performance assessment, examiner judgments may still be affected by contextual factors, examiner examinee interactions, and differences among examiners, indicating that standardization does not fully eliminate scoring variability~\cite{chong2017sights}. Third, examiner severity may vary over the course of an examination. Temporal variation in OSCE scoring has been observed, including differential rater function over time and greater leniency at the beginning of an examination~\citep{mclaughlin2009effect, hope2015examiners}. This suggests that examiner severity may shift systematically even within the same assessment session, introducing an additional source of rating variability beyond case content or candidate performance alone.

\section{Physician-Defined Scoring Rubric}
\label{Appendix_Rubric}
\begin{figure*}[t]
  \centering
  \includegraphics[width=\textwidth]{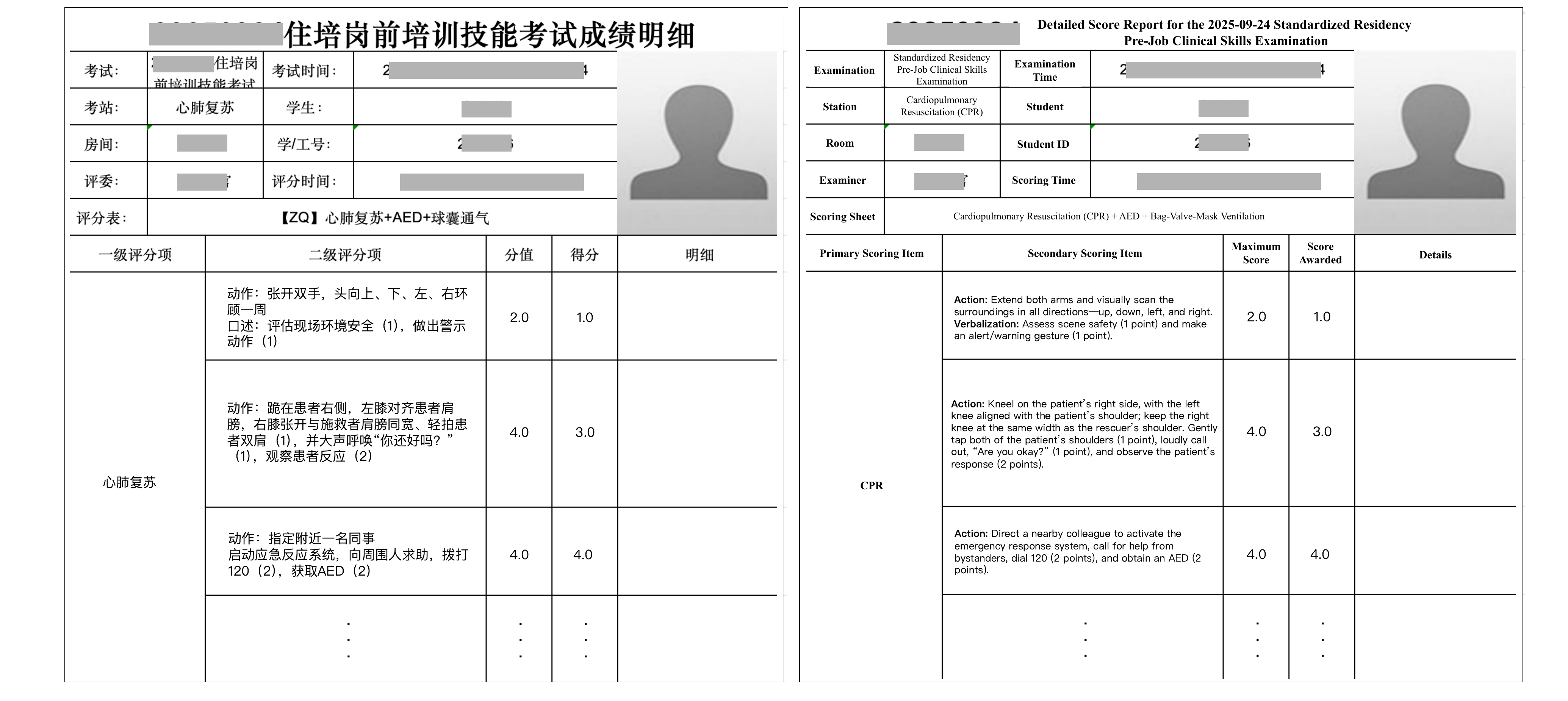}
  \caption{Representative example of the physician-defined scoring rubric in SiMing-Bench. The left panel presents the original Chinese examination form, and the right panel presents its English-translated version.}
  \label{fig:rubric}
\end{figure*}
\begin{figure*}[t]
  \centering
  \includegraphics[width=\textwidth]{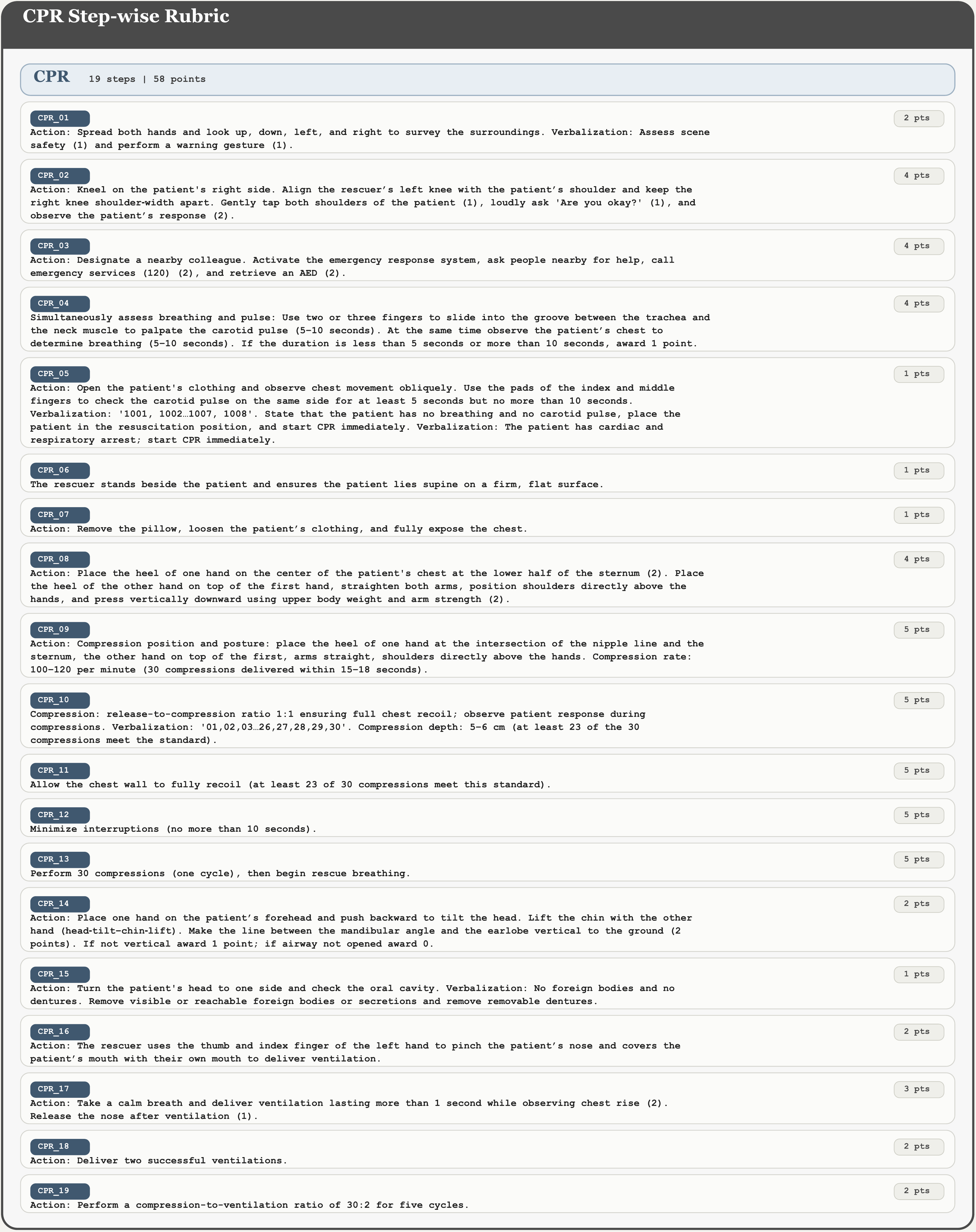}
  \caption{Standardized CPR rubric defined by physicians and used for step-level annotation in our benchmark.}
  \label{fig:bls_steps_rubric_cpr}
\end{figure*}
\begin{figure*}[t]
  \centering
  \includegraphics[width=\textwidth]{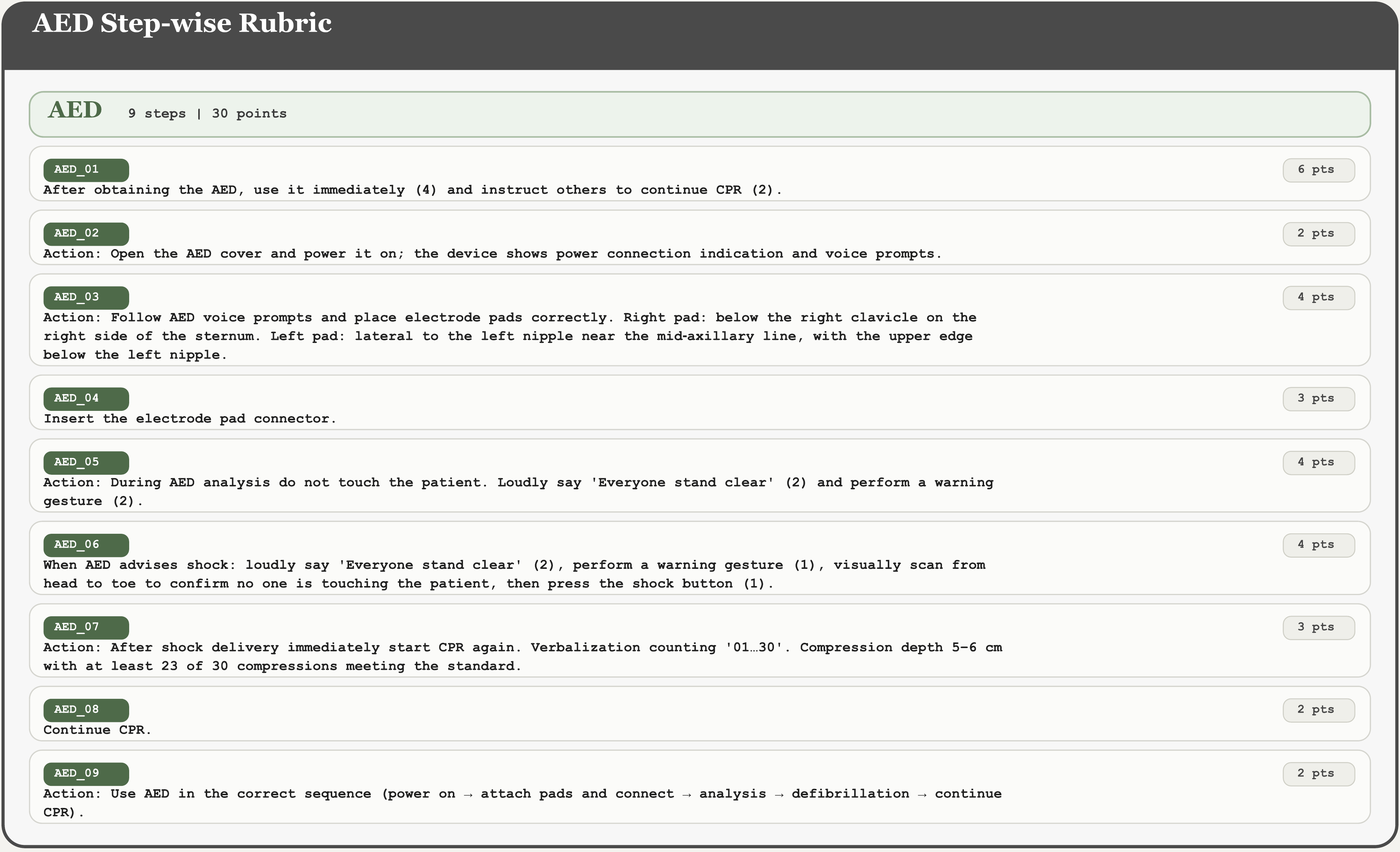}
  \caption{Standardized AED rubric defined by physicians and used for step-level annotation in our benchmark.}
  \label{fig:bls_steps_rubric_aed}
\end{figure*}
\begin{figure*}[t]
  \centering
  \includegraphics[width=\textwidth]{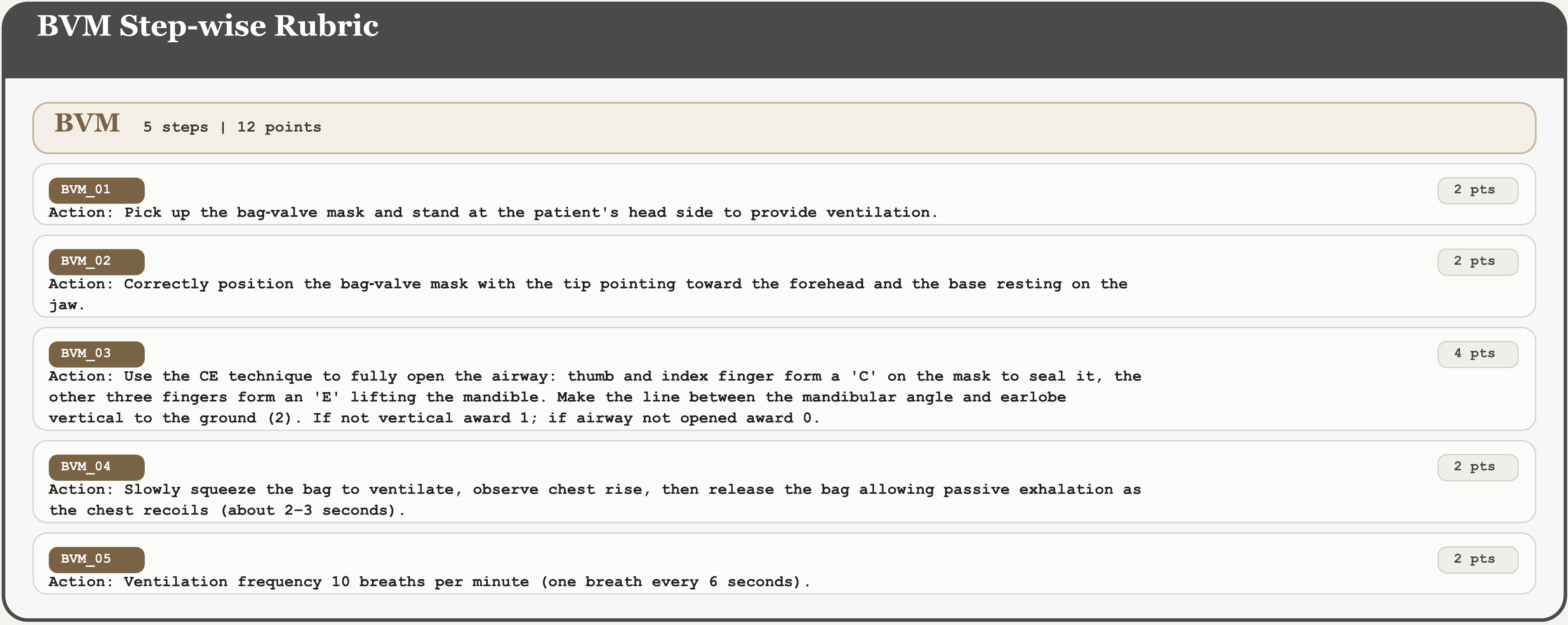}
  \caption{Standardized BMV rubric defined by physicians and used for step-level annotation in our benchmark.}
  \label{fig:bls_steps_rubric_bvm}
\end{figure*}

Each clinical skill category is paired with an official physician-defined scoring rubric used in standardized clinical skill examinations. As shown in Figure~\ref{fig:rubric}, the rubric decomposes the full procedure into clinically meaningful sub-steps and specifies, for each step, the maximum score and the scoring criteria. The resulting annotations are therefore rubric-grounded judgments of procedural correctness rather than free-form descriptions.
Across the three skill categories, the benchmark contains 33 rubric-defined steps in total: 19 for CPR, 9 for AED, and 5 for BMV. The complete set of rubric items is shown in Figure~\ref{fig:bls_steps_rubric_cpr}, Figure~\ref{fig:bls_steps_rubric_aed} and Figure~\ref{fig:bls_steps_rubric_bvm}. As reported in the main paper, each video contains the complete workflow covering all three skills. The step-level labels are assigned independently by two clinically trained physician annotators according to the rubric and then adjudicated when disagreements occur.

\section{Privacy and Annotation}
\label{appendix:Privacy and Annotation}
The student videos were collected in the context of routine clinical skills training and assessment rather than through open recruitment. During data collection, students wore surgical caps and masks throughout video recording, and facial regions in the collected videos were additionally masked to further protect identity information. In addition, personally identifying information in the scoring forms, including names, institutional affiliations, and student ID numbers, was masked. 

Both annotators are physicians from the collaborating hospital involved in data collection, with substantial experience in clinical teaching and skills examination.  Prior to formal annotation, they jointly participated in the development and refinement of the rubric. All videos were annotated independently. No crowdsourcing platform was used.
\section{Prompts and Experimental Settings}
\label{Appendix_Prompt}
\begin{figure*}[t]
  \centering
  \includegraphics[width=\textwidth]{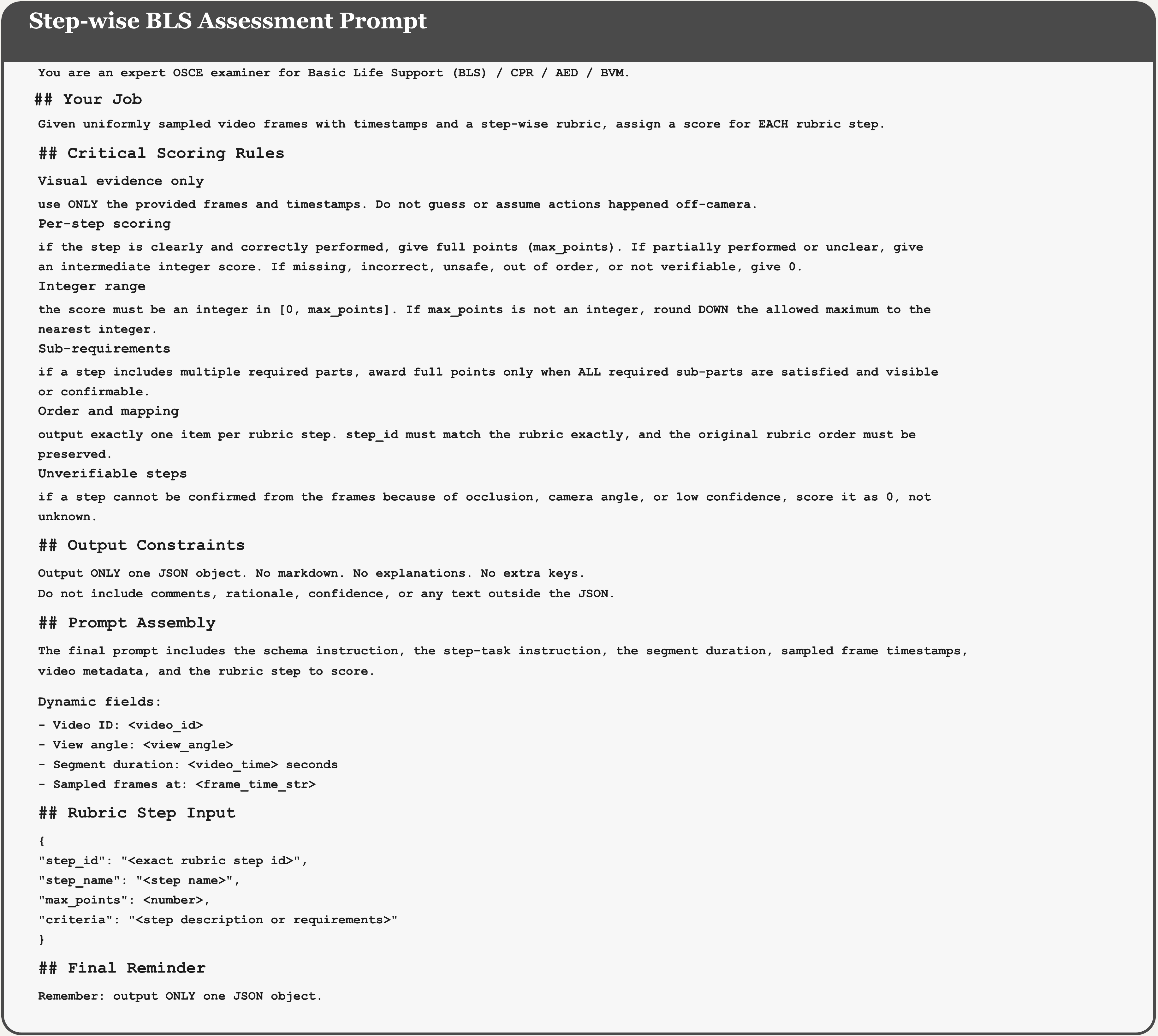}
  \caption{Prompt template for full-video step-wise scoring.}
  \label{fig:Step-wise_Prompt}
\end{figure*}

\begin{figure*}[t]
  \centering
  \includegraphics[width=\textwidth]{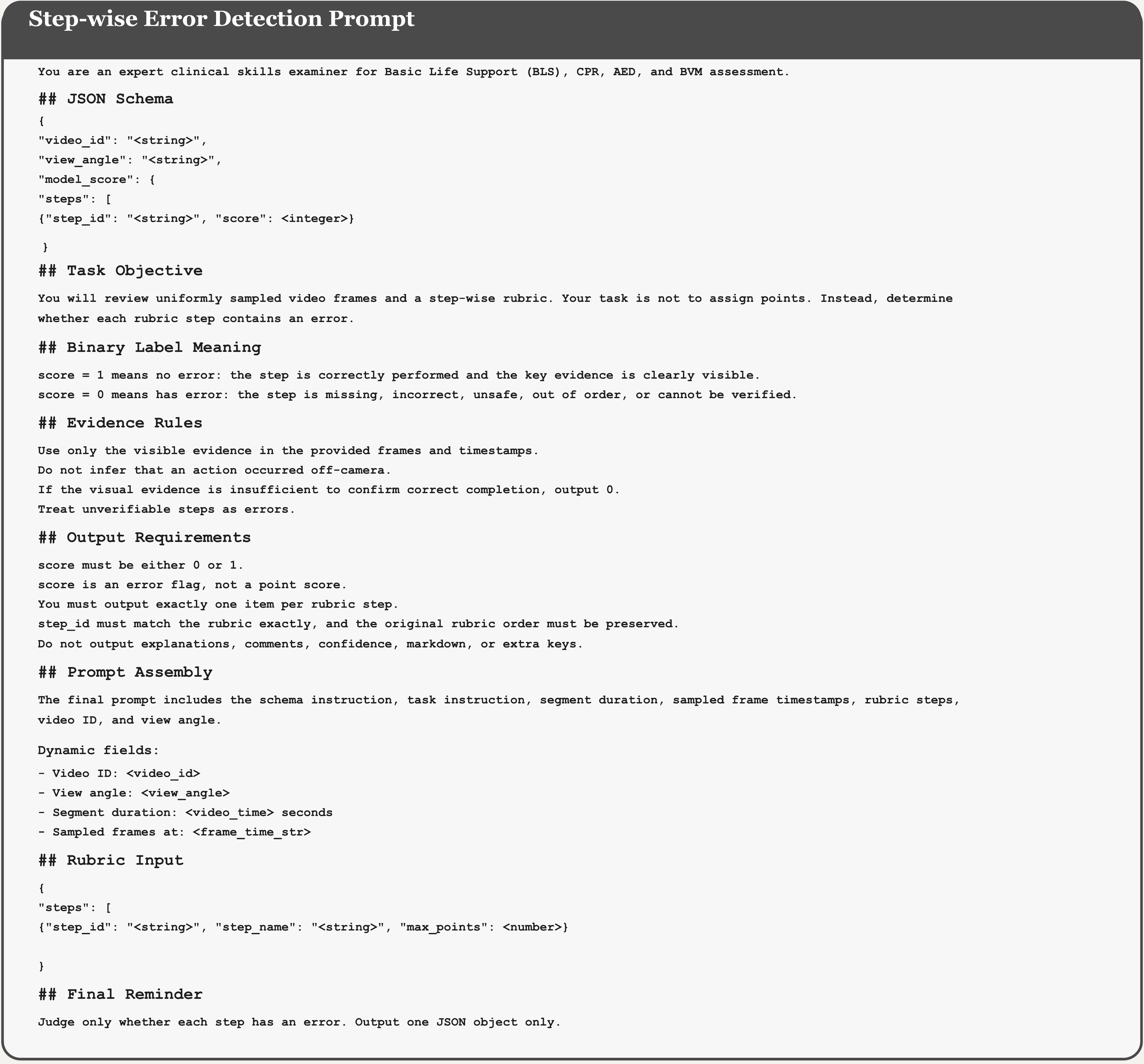}
  \caption{Prompt template for full-video step-wise error detection.}
  \label{fig:Step-wise_Error}
\end{figure*}

\begin{figure*}[t]
  \centering
  \includegraphics[width=\textwidth]{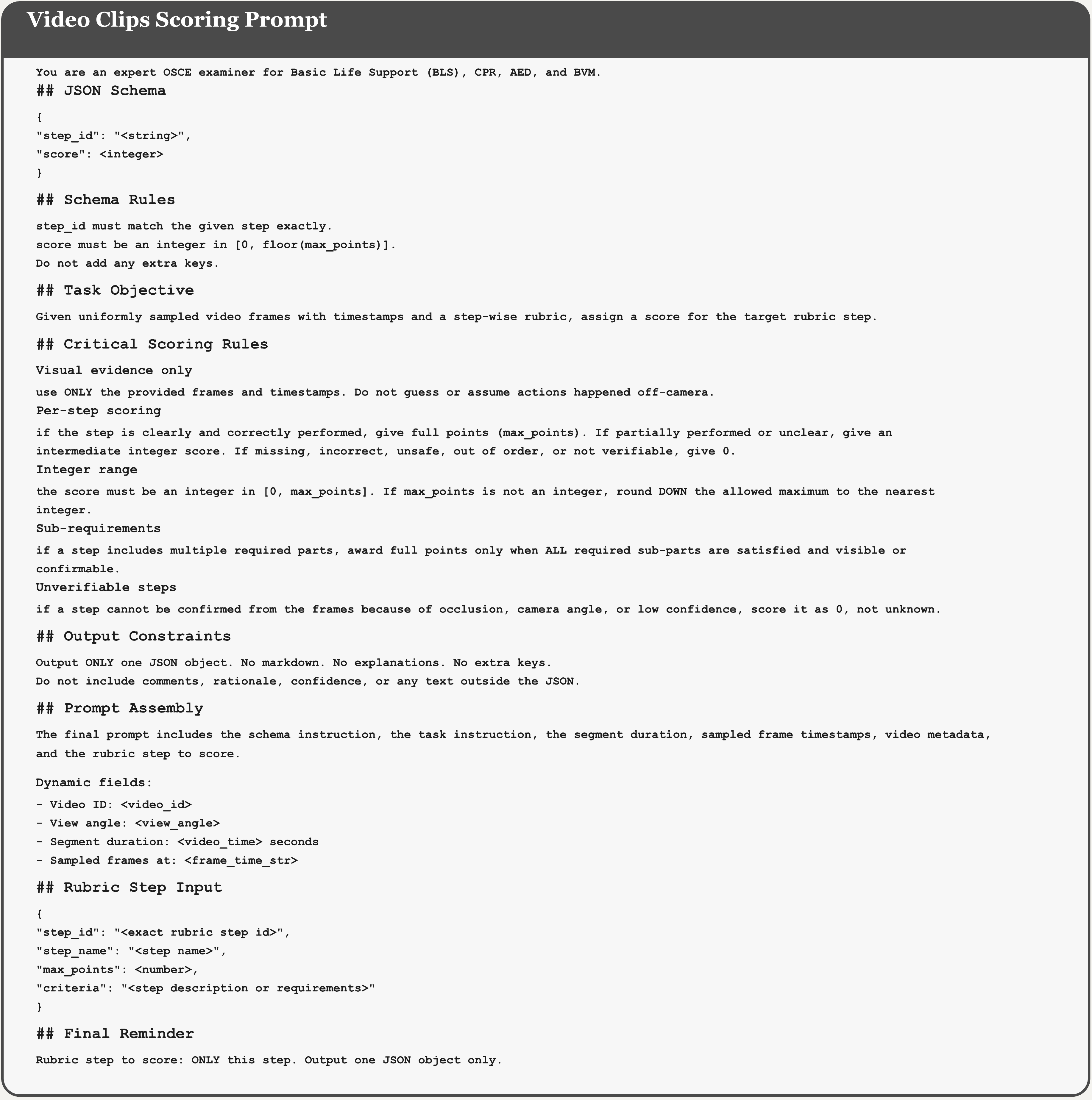}
  \caption{Prompt template for step-aligned clip scoring.}
  \label{fig:Video_Clips_Scoring_Prompt}
\end{figure*}

\begin{figure*}[t]
  \centering
  \includegraphics[width=\textwidth]{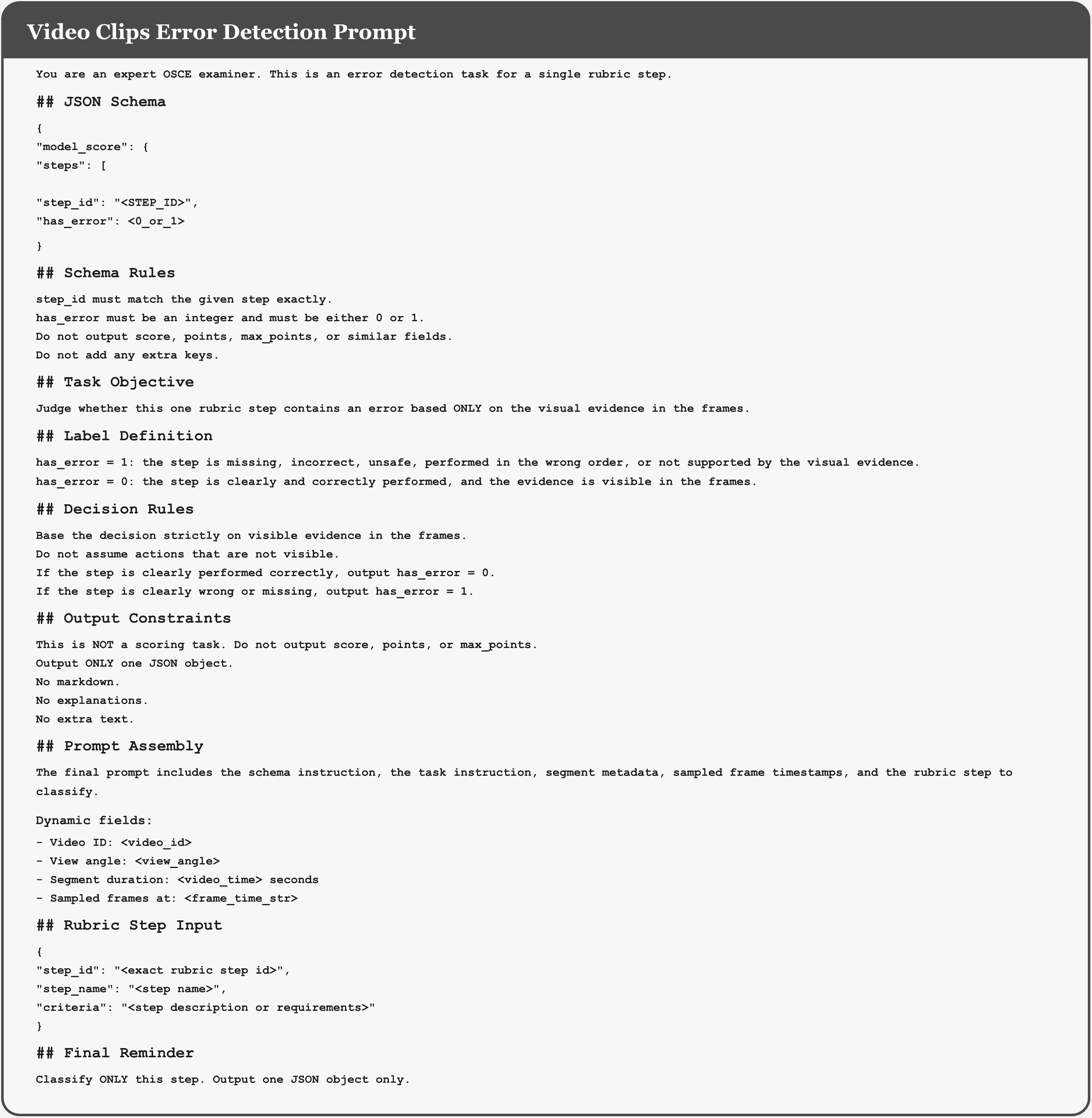}
  \caption{Prompt template for step-aligned clip error detection.}
  \label{fig:Video_Clips_Error_Detection_Prompt}
\end{figure*}

To ensure a fair and consistent comparison across all models, we employed a standardized evaluation protocol, where each model was evaluated in a single run under the same prompting setup. For each task type, we used a unified prompt template, modifying only the task-specific instructions and expected output formats when necessary. This design minimizes prompt-induced variation and helps ensure that observed performance differences are attributable to the models themselves rather than to differences in prompt construction. The four prompt templates correspond to the four evaluation settings studied in the paper: full-video step-wise scoring(Figure~\ref{fig:Step-wise_Prompt}), full-video step-wise error detection(Figure~\ref{fig:Step-wise_Error}), step-aligned clip scoring(Figure~\ref{fig:Video_Clips_Scoring_Prompt}), and step-aligned clip error detection(Figure~\ref{fig:Video_Clips_Error_Detection_Prompt}). Taken together, these four templates form a controlled prompt suite for the main and diagnostic experiments, enabling a more interpretable analysis of whether current MLLMs fail because of scoring granularity, temporal localization, or deeper limitations in converting continuous interaction processes into expert-aligned procedural assessments.

For each model, we used its official default inference settings. For models that do not natively support video input, we uniformly sampled frames from each video according to the maximum number of frames allowed by the model. The number of input frames used for each model is reported in Table~\ref{Appendix:settings}.

\begin{table}[t]
\centering
\small
\setlength{\tabcolsep}{6pt}
\begin{tabular}{lc}
\toprule
Models & Frames \\
\midrule
LLaVA-Video & 64 \\
LLaVA-OneVision & 64 \\
Qwen3-VL & 64 \\
Qwen3-Omni & 60 \\
GPT-4o & 50 \\
UniMed-VL & 30 \\
MedGemma & 30 \\
MediX-R1 & 30 \\
Doubao-1.6-Vision & 240 \\
Gemini-3 Pro & 120 \\
GPT-5.2 & 50 \\
\bottomrule
\end{tabular}
\caption{The evaluated models and the number of input frames used for each model.}
\label{Appendix:settings}
\end{table}

\section{Metric}
\subsection{Overall-score Consistency}
\label{Appendix_Metric_Overall-score}
Formally, given expert overall scores $\{x_i\}_{i=1}^{n}$ and model-predicted overall scores $\{y_i\}_{i=1}^{n}$, Pearson correlation is computed as
\begin{equation}
r_p =
\frac{
\sum_{i=1}^{n}(x_i-\bar{x})(y_i-\bar{y})
}{
\sqrt{\sum_{i=1}^{n}(x_i-\bar{x})^2}\,
\sqrt{\sum_{i=1}^{n}(y_i-\bar{y})^2}
}.
\end{equation}
where $\bar{x}$ and $\bar{y}$ are the sample means of expert and predicted scores, respectively.

Let $u_i=\mathrm{rank}(x_i)$ and $v_i=\mathrm{rank}(y_i)$ denote the ranks of expert and predicted overall scores. Spearman correlation is then computed as
\begin{equation}
r_s =
\frac{
\sum_{i=1}^{n}(u_i-\bar{u})(v_i-\bar{v})
}{
\sqrt{\sum_{i=1}^{n}(u_i-\bar{u})^2}\,
\sqrt{\sum_{i=1}^{n}(v_i-\bar{v})^2}
},
\end{equation}
where $\bar{u}$ and $\bar{v}$ are the mean ranks of $\{u_i\}_{i=1}^{n}$ and $\{v_i\}_{i=1}^{n}$, respectively.

\begin{figure*}[t]
  \centering
  \includegraphics[width=\textwidth]{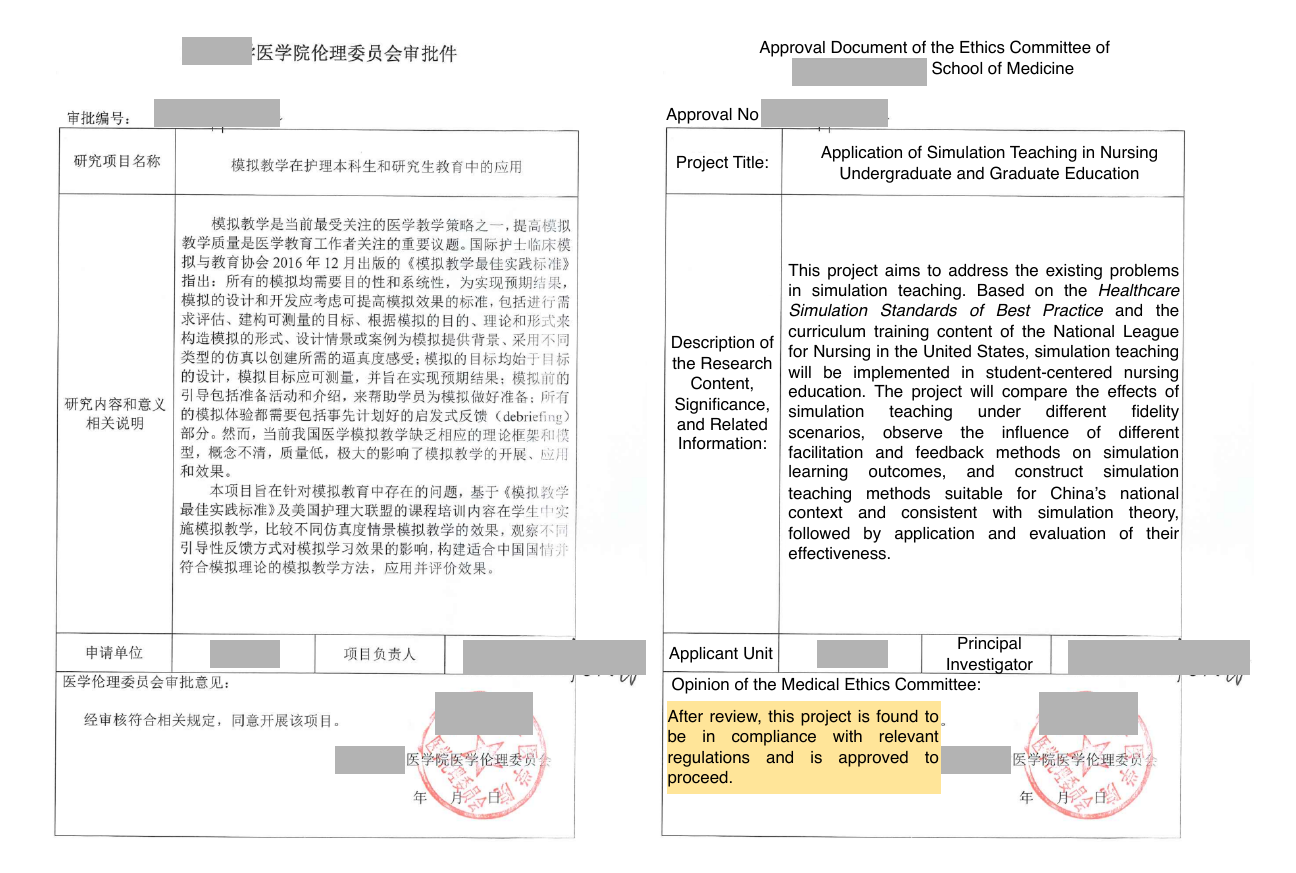}
  \caption{Screenshot of the IRB approval document.}
  \label{fig:ethic}
\end{figure*}

\subsection{Step-level scoring consistency}
\label{Appendix_Metric_Step-score}

To improve the interpretability of the score-to-category conversion, we adopt a \textbf{rubric-aligned mapping scheme}. Let $s_{\max}$ denote the maximum score for a procedural step, $d$ the number of deducted points, and $s = s_{\max} - d$ the resulting raw score. The ordinal label $g(s)$ is defined as
\begin{equation}
g(s)=
\begin{cases}
3, & d = 0,\\
2, & d = 1,\\
1, & 1 < d < s_{\max},\\
0, & s = 0.
\end{cases}
\end{equation}
This mapping follows the logic of the original clinical rubric: label $3$ corresponds to fully correct execution, label $2$ to execution with only a minor deduction, label $1$ to clearly deficient execution with partial credit retained, and label $0$ to severely incorrect or failed execution. Given these ordinal labels, we compute \textbf{quadratically weighted Cohen's $\kappa$} to measure agreement between expert annotations and model predictions. Let $O_{ij}$ denote the observed proportion of samples assigned to expert label $i$ and model label $j$, and let $E_{ij}$ denote the corresponding expected proportion under statistical independence. With $K=4$ ordinal categories, the quadratic weight is defined as
\begin{equation}
w_{ij}=\frac{(i-j)^2}{(K-1)^2}.
\end{equation}
The weighted observed disagreement and expected disagreement are given by
\begin{equation}
D_o =
\sum_{i=0}^{K-1}\sum_{j=0}^{K-1}
w_{ij} O_{ij},
\end{equation}
and
\begin{equation}
D_e =
\sum_{i=0}^{K-1}\sum_{j=0}^{K-1}
w_{ij} E_{ij}.
\end{equation}
The final quadratically weighted Cohen's $\kappa$ is computed as
\begin{equation}
\kappa = 1 - \frac{D_o}{D_e}.
\end{equation}
A higher $\kappa$ indicates stronger agreement between model predictions and expert ratings on ordered step-level judgments, while assigning larger penalties to disagreements that are farther apart on the ordinal scale.

\section{Extended Ethical Considerations}
\label{Appendix_Ethical}
This work uses video data collected from real clinical skill examinations involving medical students. The data collection and research use protocols were reviewed and approved by the Institutional Review Board (IRB) of the participating institution. All data were collected under institutional oversight in standard clinical training and examination settings and were used solely for education, training, and research purposes. All data were handled under approved ethical procedures, with appropriate measures for privacy protection, controlled access, and research-only use. A screenshot of the IRB approval document is provided in Figure~\ref{fig:ethic}.

\end{document}